\pdfoutput=1
\documentclass[letterpaper, 10 pt, conference]{ieeeconf}  

\IEEEoverridecommandlockouts                              

\usepackage{graphicx}
\usepackage{color}
\usepackage{comment}
\usepackage{url}
\usepackage{multirow}
\usepackage{placeins}
\usepackage{makecell}
\usepackage{caption}
\usepackage{booktabs, siunitx, threeparttable, makecell}
\usepackage{amsmath, amssymb}
\usepackage{algorithm}
\usepackage{algpseudocode}

\newcommand{\minusfeat}{FRAM $-$Feat.}
\newcommand{\minustraj}{FRAM $-$Traj.}

\title{\LARGE \bf FRAM: Trajectory-Guided Visual Feature Selection for\\ Compact Language-Conditioned Robot Manipulation}
\author{Hiroshi Ito$^{1}$, Hyogo Hiruma$^{1}$, Yoshiki Kanai$^{1}$, Takahiro Yoshida$^{1}$, Akira Kanazawa$^{1}$, Hiroyuki Yamada$^{1}$
\thanks{$^{1}$Research \& Development Group, Hitachi, Ltd., Ibaraki, 312-0034, Japan.
{\tt\small hiroshi.ito.ws@hitachi.com}}
}

\begin{document}

\maketitle
\thispagestyle{empty}
\pagestyle{empty}

\begin{abstract}
Vision-Language-Action (VLA) models achieve strong performance in robot manipulation, but often require large numbers of parameters.
In this work, we propose the Future Representation Action Model (FRAM), a small policy that explicitly links the future end-effector trajectory to the current visual input.
FRAM uses the image coordinates of the predicted trajectory as spatial pointers and reads local visual features related to the motion from the current image.
This organizes the information for action generation into the reference position (Where), the visual state (What), and the future motion (Future).
Trajectory labels are generated automatically from demonstrations and camera geometry, so no manual annotation is needed.
With 138.7M parameters, including a frozen language encoder, FRAM reaches an average success rate of 92.2\% over the four standard LIBERO suites, close to the 94.2\% of $\pi_0$ with 3.3B parameters.
Without extra training, it also reaches an average of 67.3\% on LIBERO-Plus.
Ablations confirm that both the future trajectory and the local visual features improve performance and robustness.
On a real dual-arm UR5e, FRAM stacks cups using only wrist cameras, including choosing and switching between the left and right arms.
These results show that selecting visual information based on future motion is an effective way to obtain both high performance and robustness in a small robot policy.

\end{abstract}

\section{Introduction}
Robots are moving from structured settings such as factories and warehouses into more varied sites such as maintenance and service work. In these settings, a robot must produce the right action when the objects, their layout, or the language instruction change. Recently, language-conditioned policies such as Vision-Language-Action (VLA) models have enabled a wide range of robot manipulation from images and language instructions. In particular, models with billions of parameters, such as OpenVLA~\cite{openvla} and $\pi_0$~\cite{pi0}, learn the relation between vision, language, and action from large and diverse data. However, fine-tuning and inference for each site require large computing resources. To use such policies where computing resources are limited, we need small models that still perform well.

For a small model to perform well, it must select the visual information needed for the next action and represent it in a form that a limited model capacity can use. Visuomotor learning with spatial attention extracts image positions related to the action as a low-dimensional intermediate representation and learns the mapping from vision to action~\cite{spatialae}. However, image coordinates alone do not keep enough information about the appearance of the object and its surroundings, and the current attention position alone cannot explicitly represent the future motion path and the visual information related to it. Trajectory representations extend image points along time. ATM~\cite{atm} combines future trajectories learned from videos with image features to generate actions, MT-$\pi$~\cite{mtpi} uses 2D end-effector trajectories as the action representation, and RT-Trajectory~\cite{rttrajectory} conditions action generation on a trajectory sketch given from outside. These methods use trajectories as motion predictions, action representations, or action instructions, but they do not explicitly read the needed visual information from the current image along the predicted trajectory. As a result, the relation between the future path and the objects and surroundings along that path is learned only implicitly inside the model.

In this work, we propose the Future Representation Action Model (FRAM), which links the information needed for action generation as the reference position (Where), the current visual state (What), and the future motion (Future). FRAM is inspired by the finding that people look at the next object or target location before moving the hand~\cite{land1999,johansson2001}. Using pretrained vision and language models, FRAM predicts the future end-effector trajectory as a sequence of points with image coordinates, depth, and projected orientation from two RGB views, the end-effector state, and the language instruction. The predicted trajectory is not only a condition for action generation. Its image coordinates also serve as spatial pointers into the feature map of the current image, from which FRAM reads local features along the trajectory and, using the end point of the prediction as a query, the context of the target location from the whole image. These local features and the trajectory representation are combined with global image features, so the policy keeps information about the whole scene while focusing action generation on the information related to the predicted motion. The continuous action sequence of the end-effector is generated by flow matching~\cite{flowmatching}. Trajectory labels are generated automatically from the end-effector poses, gripper commands, and camera geometry recorded in the demonstrations, so no manual annotation of object regions or subgoals is needed. The contributions of this paper are threefold.

\textbf{Structuring visual information with future trajectories}: We propose a predictive intermediate representation that links position, state, and future. The predicted end-effector trajectory is used as spatial pointers, and the local features along the trajectory and the context of the target location are connected to action generation.

\textbf{High performance and robustness with a small policy}: With 138.7M parameters, including a frozen language encoder, FRAM reaches an average success rate of 92.2\% over the four standard LIBERO suites, close to the 94.2\% of $\pi_0$~\cite{oft}, which has 3.3B parameters and is about 24 times larger.
Without extra training on perturbed environments, it reaches an average of 67.3\% on LIBERO-Plus. In ablations, removing the future trajectory or the local visual features lowers performance, which confirms that both contribute to performance and to robustness against environment changes.

\textbf{Application to real dual-arm manipulation}: In a cup stacking task with a dual-arm UR5e, using only wrist cameras, the predicted trajectory is updated step by step according to the target object and the placement location, and the policy generates a sequence of operations that includes choosing and switching between the left and right arms.

\section{Related Work}
\subsection{Efficient and Compact VLA}
\label{ssec:vla}

In VLA, model structure, action representation, and training methods have been studied to connect pretrained vision and language knowledge to robot actions efficiently. OpenVLA~\cite{openvla} trains a pretrained vision-language model on robot demonstrations and generates action tokens from images and language instructions. $\pi_0$~\cite{pi0} combines a vision-language model with an action expert based on flow matching to generate continuous action sequences. OpenVLA-OFT~\cite{oft} adds parallel decoding, action chunking, and continuous action representations, and RIPT-VLA~\cite{ript} improves the policy by post-training with environment interaction.
To reduce computation, SmolVLA~\cite{smolvla} combines a small vision-language model with a flow matching action generator, and FLOWER~\cite{flower} reduces model size by rethinking how capacity is split between the vision-language part and the action generator and how they are connected. Octo~\cite{octo} processes vision and language in separate modules, which allows adaptation to different observation and action spaces. These studies show that network size, capacity allocation across modules, and the design of the action generator strongly affect policy performance and computational efficiency.

For a small model to perform well, however, the representation of the information given to the action generator also matters, not only the model structure. In particular, rather than using features uniformly from the whole image, the policy should select the visual information related to the motion it is about to execute and structure it in a form that a limited model capacity can use.

\subsection{Action Generation via Trajectory and Future Representations}
\label{ssec:traj}

One way to treat future motion as an explicit intermediate representation is to represent the movement of image points as trajectories. ATM~\cite{atm} learns point trajectories from videos without action labels and combines future trajectories predicted from images and language with image features for action generation. MT-$\pi$~\cite{mtpi} uses 2D motion tracks shared between humans and robots as the action representation and converts trajectories predicted from multiple views into 3D end-effector motion geometrically. RT-Trajectory~\cite{rttrajectory} encodes a trajectory sketch given by a human or a high-level module together with the image and conditions action generation on it. These studies show that trajectories work well as future motion predictions, as a shared action representation, or as action instructions.
Other work treats the future state as a higher-dimensional representation. CoT-VLA~\cite{cotvla} generates a future subgoal image as a condition for action generation, and ThinkAct~\cite{thinkact} feeds a visual latent plan to a low-level action generator. Like trajectory representations, these methods help action generation by making information about the future explicit.

In these methods, however, the predicted trajectory or future representation is mainly used as a condition for action generation or as an action representation. Making explicit which region of the current image the trajectory corresponds to, and directly reading the visual state needed for the action from the path and the target location, has not been fully explored. FRAM uses the image coordinates of the predicted end-effector trajectory as spatial pointers into the feature map of the current image. By explicitly linking the future motion representation to the current visual information, it selectively uses the information related to the action.

\section{Proposed Method}
\begin{figure*}[t]
\centering
\includegraphics[width=0.98\textwidth]{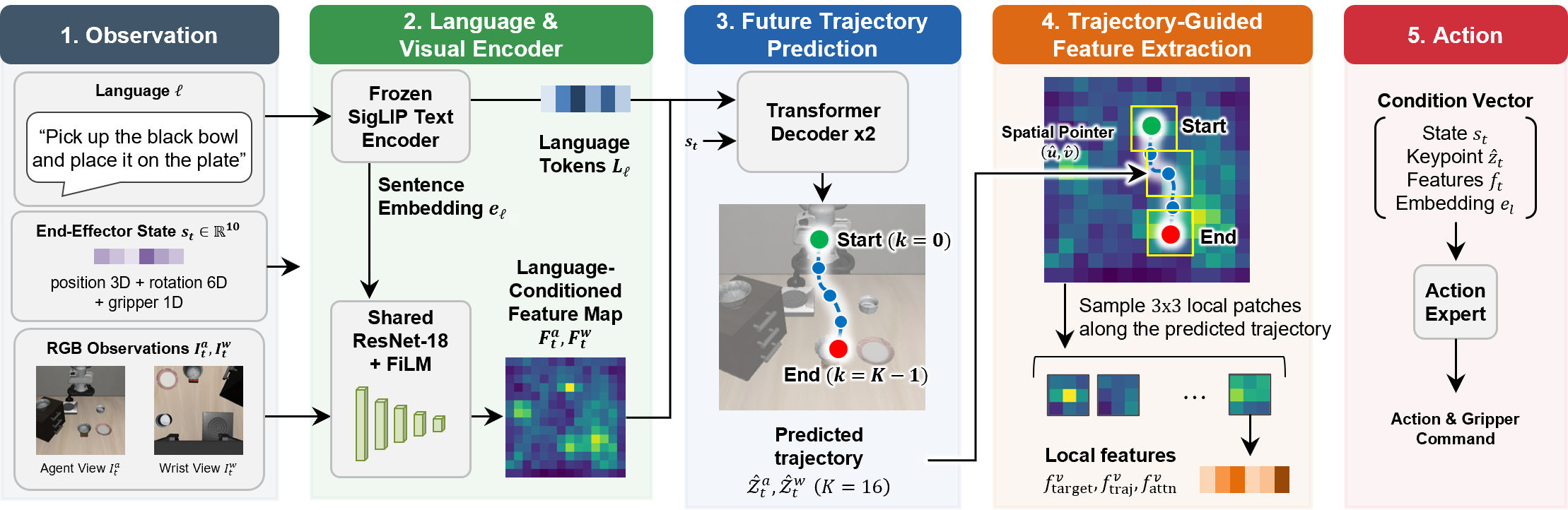}
\caption{Overview of FRAM. (1) Observation inputs, (2) language and vision encoders, (3) prediction of the future end-effector trajectory, (4) trajectory-guided local feature extraction, and (5) action generation by flow matching.}
\label{fig:model_overview}
\end{figure*}

\subsection{Design Concept}
\label{ssec:overview}

In object manipulation, people look at the next object or target location before moving the hand~\cite{johansson2001}. Inspired by this, FRAM predicts the future end-effector trajectory in image coordinates from the current observation and uses these coordinates as spatial pointers to read local features related to the action from the feature map of the current image. The predicted trajectory represents where to look (Where) and where to move (Future), and the features read from those positions represent the state of the object and its surroundings (What). Unlike a policy that generates actions only from a global feature that compresses the whole image, the image positions the policy refers to are geometrically tied to the future motion. This makes it harder for changes in the background or in surrounding objects unrelated to the task to leak into the condition.

Fig.~\ref{fig:model_overview} shows the structure of FRAM. (1) The observation at time $t$ consists of the language instruction $\ell$, the end-effector state $s_t$, the external camera image $I_t^a$, and the wrist camera image $I_t^w$. (2) The language and vision encoders produce language-conditioned feature maps $F_t^v$ ($v\in\{a,w\}$). (3) The future trajectory predictor predicts a $K$-point future end-effector trajectory $\hat{\mathcal Z}_t^v$ for each view from $F_t^v$, the language, and the end-effector state. (4) Trajectory-guided feature extraction reads local features from $F_t^v$ using the image coordinates of the predicted trajectory as spatial pointers. (5) Finally, the action expert generates an $H$-step action chunk by flow matching, conditioned on a vector made of the end-effector state, the predicted trajectory, the local features, and the language embedding.

\subsection{Language and Vision Encoders}
\label{ssec:encoder}

The language instruction $\ell$ is encoded by a frozen SigLIP text encoder~\cite{siglip}, giving a token sequence $L_\ell$ and a sentence embedding $e_\ell$. Images are encoded by an ImageNet-pretrained ResNet-18 up to its intermediate stage (the third residual stage), with weights shared between the two views. The feature map $\phi(I_t^v)$ is projected to a lower dimension by a $1\times1$ convolution, and then FiLM~\cite{film} with the scale $\gamma(e_\ell)$ and shift $\beta(e_\ell)$ generated from $e_\ell$,
\begin{equation}
F_t^v=\bigl(1+\gamma(e_\ell)\bigr)\odot\phi(I_t^v)+\beta(e_\ell),
\label{eq:film}
\end{equation}
is applied at each spatial position to obtain the language-conditioned feature map $F_t^v$. With FiLM, the regions that are emphasized change with the instruction even for the same image.
The end-effector state $s_t$ is the concatenation of the end-effector position, a 6D continuous rotation representation~\cite{rot6d}, and the gripper opening, and is embedded by an MLP. The global visual features of both views, the end-effector state embedding, and the language embedding are concatenated and passed through an MLP to give the global observation feature $g_t$. $g_t$ summarizes the whole image and is used as a condition for action generation together with the local features described below.

\subsection{Future Trajectory Prediction and Label Generation}
\label{ssec:kpt}

\textbf{Prediction.} A separate Transformer decoder is used for each view. $K$ learnable queries cross-attend to a memory made of image tokens (the flattened $F_t^v$ with position embeddings), the language tokens $L_\ell$, and an end-effector state token, and directly regress the future trajectory. The $k$-th predicted point is
\begin{equation}
\hat{\mathbf z}_{t,k}^{v}
=
\begin{bmatrix}
(\hat{\mathbf q}_{t,k}^{v})^\top &
\hat d_{t,k}^{v} &
(\hat{\mathbf o}_{t,k}^{v})^\top
\end{bmatrix}^{\top},
\label{eq:traj_repr}
\end{equation}
where $\hat{\mathbf q}_{t,k}^{v}=[\hat u,\hat v]^\top$ is the normalized image coordinate, $\hat d_{t,k}^{v}$ is the depth in the camera frame, and $\hat{\mathbf o}_{t,k}^{v}\in\mathbb{R}^{6}$ is the orientation offset obtained by projecting the three end-effector axes onto the image.

\textbf{Label generation.} Trajectory labels are generated automatically from the end-effector pose sequence, the gripper commands, and the camera geometry of each demonstration. First, the demonstration is split into subtasks (approach, grasp, transport, release, etc.) at the times when the gripper command switches. The end $e(t)$ of the prediction interval at time $t$ is the later of the end of the subtask containing $t$ and $t+H$, cut off at the end of the demonstration. The predicted trajectory therefore heads toward the end of the current subtask, and the interval does not shrink below $H$ steps just before the end of a subtask. Next, a piecewise cubic B-spline is fitted in the base frame to the end-effector positions in $[t,e(t)]$. The current position, the end of the interval, and the positions where the gripper switches within the interval are knots that the curve passes through exactly. $K$ points are then sampled from the curve at equal arc-length intervals. This gives a point sequence that represents the geometry of the path regardless of speed variations in the demonstration. The first point $k=0$ is the current end-effector position, and the last point $k=K-1$ is the end of the interval. The orientation of each point is obtained by interpolating the end-effector orientation at the corresponding time.

Finally, the $K$ positions and end-effector axes are projected onto the image of each view using the camera geometry at the current time $t$. A point ${}^{b}\mathbf p_k$ in the base frame is transformed to the camera frame by the camera extrinsics ${}^{v}\mathbf T_b^{(t)}$ at time $t$ and projected by the intrinsic matrix $\mathbf K_v$ as
\begin{equation}
\begin{bmatrix}
{}^{v}\mathbf p_{k}^{(t)}\\
1
\end{bmatrix}
=
{}^{v}\mathbf T_b^{(t)}
\begin{bmatrix}
{}^{b}\mathbf p_{k}\\
1
\end{bmatrix},
\qquad
\mathbf q_{t,k}^{v}
=
\Pi\!\left(
\mathbf K_v\,{}^{v}\mathbf p_{k}^{(t)}
\right),
\label{eq:projection}
\end{equation}
where $\Pi([x,y,z]^\top)=[x/z,y/z]^\top$. For the external camera, calibrated extrinsics and intrinsics are used. For the wrist camera, ${}^{v}\mathbf T_b^{(t)}$ is computed from the current end-effector pose and the hand-eye transform. Pixel coordinates are normalized to $[-1,1]$, and the depth is the distance along the camera axis. The orientation label is obtained by projecting the end-effector origin and points on its three axes in the same way and concatenating the image-space differences between each axis tip and the origin into a 6D offset. Since all future end-effector poses are projected into the camera frame at the current time, no future camera poses are needed even for the wrist camera, and the predicted coordinates can be used directly as reference positions on the feature map of the current image.

\textbf{Points outside the image or behind the camera.} For points projected outside the image, the true projected coordinates outside $[-1,1]$ are used as labels, which keeps the information that the motion heads out of view. Points behind the camera ($z\le0$) and points far from the optical axis, where $1/z$ in Eq.~\eqref{eq:projection} nearly diverges, are excluded from the loss.

\subsection{Trajectory-Guided Image Feature Extraction}
\label{ssec:local}

Using the image coordinates $\hat{\mathbf q}_{t,k}^{v}$ of the predicted trajectory as spatial pointers, features related to the action are read from the feature map $F_t^v$ of the current image (Fig.~\ref{fig:model_overview}(4)). Each predicted point is mapped to grid coordinates on $F_t^v$, a neighborhood grid centered at that point ($3\times3$ in the figure) is sampled by bilinear interpolation, and the uniform average is taken as $h_{t,k}^v$. The sampling is implemented with a differentiable grid sample. Coordinates outside the image are clipped to the edge of the feature map, and the edge values are used. Only the image coordinates are used for reading; the depth and orientation offset are passed to the condition vector later.

The average over the $K$ points, $f_{\mathrm{traj}}^v=\frac{1}{K}\sum_k h_{t,k}^v$, represents the visual state along the path (the grasped object and nearby objects along the path), and the end-point feature $f_{\mathrm{target}}^v=h_{t,K-1}^v$ represents the visual state of the target location (the next object to grasp or the placement location). In addition, a single attention layer with $f_{\mathrm{target}}^v$ as the query and all tokens of $F_t^v$ as keys and values gives the context feature $f_{\mathrm{attn}}^v$. This term fills in features from regions similar to the end-point feature when the predicted end point is slightly off the target. The three features are concatenated and combined into the local visual feature as
\begin{equation}
f_{\mathrm{local}}^v
=
\operatorname{MLP}
\left(
[
f_{\mathrm{traj}}^v;
f_{\mathrm{target}}^v;
f_{\mathrm{attn}}^v
]
\right).
\label{eq:local_feature}
\end{equation}
This is done independently for each view. Therefore, What in FRAM is not a prediction of a future image but the visual state in the current image selected by the predicted future motion.

\subsection{Training and Generation with the Action Expert}
\label{ssec:action}

\textbf{Condition.} The condition vector $c_t$ for action generation is the concatenation of the global observation feature $g_t$, a trajectory embedding obtained by flattening the predicted trajectory $\hat{\mathcal Z}_t^v$ of each view and passing it through an MLP, and the local visual feature $f_{\mathrm{local}}^v$ of each view (Fig.~\ref{fig:model_overview}(5)). In addition, each predicted point $\hat{\mathbf z}_{t,k}^{v}$ is converted into a separate trajectory token by a linear projection and a learnable position embedding, and these tokens are given to the action generator as cross-attention memory. This lets each step of the action chunk directly refer to the position, depth, and orientation of each point on the trajectory.

\textbf{Action generation by flow matching.} Let the action chunk be $A_1\in\mathbb{R}^{H\times d_a}$. Each step consists of delta translation and rotation commands for the end-effector and a gripper command. The gripper command is also generated by the same flow as a one-dimensional continuous value in $\pm1$, and its sign decides open or close at execution time~\cite{pi0}. In training, we sample $A_0\sim\mathcal{N}(0,I)$ and $\tau\sim\mathcal{U}(0,1)$ and form $A_\tau=(1-\tau)A_0+\tau A_1$. The velocity field $v_\theta$ treats each step of the chunk as a token with a temporal position embedding. Each block applies, in order, (i) a residual MLP with FiLM conditioning from the embeddings of $c_t$ and $\tau$ and the language embedding, and (ii) a Transformer layer with self-attention across steps and cross-attention to the trajectory tokens. The loss is
\begin{equation}
\mathcal{L}_{\mathrm{FM}}
=
\mathbb{E}
\left[
\left\|
v_\theta(A_\tau,\tau,c_t,e_\ell,\hat{\mathcal Z}_t)
-
(A_1-A_0)
\right\|_{\mathrm{valid}}^2
\right]
\label{eq:fm}
\end{equation}
following~\cite{flowmatching}, where $\|\cdot\|_{\mathrm{valid}}^2$ is the mean squared error over the valid elements, excluding the padding after the end of the demonstration.

\textbf{Total loss.} A smooth-$L_1$ loss is used for the image coordinates and depth of the trajectory. The second-order differences over three adjacent points are also matched to the labels so that the model learns the shape of the trajectory. A smooth-$L_1$ loss is also used for the orientation offset. In all cases, the excluded points described above are masked out before averaging. The total loss is
\begin{equation}
\mathcal{L}
=
\mathcal{L}_{\mathrm{FM}}
+
\mathcal{L}_{\mathrm{BC}}
+
\lambda_{\mathrm{KP}}\mathcal{L}_{\mathrm{KP}}
+
\lambda_{\mathrm{ORI}}\mathcal{L}_{\mathrm{ORI}},
\label{eq:loss}
\end{equation}
where $\mathcal{L}_{\mathrm{BC}}$ is an auxiliary loss that directly regresses the action chunk from $c_t$, and $\mathcal{L}_{\mathrm{KP}}$ and $\mathcal{L}_{\mathrm{ORI}}$ are the losses on the position and shape of the trajectory and on the projected orientation, respectively. During training, feature extraction and action generation also use the predicted trajectory, not the label. However, a stop-gradient is applied to the predicted trajectory passed to later stages, so the action loss is not backpropagated to the trajectory predictor. This keeps the geometric meaning of the trajectory through the label signal, while the same predicted trajectory is used in both training and inference.

\textbf{Inference.} At inference, $F_t^v$, $\hat{\mathcal Z}_t^v$, $f_{\mathrm{local}}^v$, and $c_t$ are computed in order from the observation. $A_0$ is sampled from Gaussian noise, and the velocity field of Eq.~\eqref{eq:fm} is integrated numerically with a fixed number of steps to obtain $A_1$. After the generated $H$-step chunk is executed, a new observation is taken, and trajectory prediction and action generation are repeated.

\section{Evaluation on LIBERO Simulation}
\subsection{Setup}
\label{ssec:sim_setup}

\textbf{Benchmarks.} We evaluate the manipulation performance of the proposed method and its robustness to environment changes on LIBERO~\cite{libero} and LIBERO-Plus~\cite{liberoplus}. LIBERO is a MuJoCo-based simulation environment with a 7-DoF Franka Emika Panda manipulator with a parallel gripper, an external camera (agent view) and a wrist camera (wrist view), and objects that mimic everyday items. The standard evaluation uses four suites: LIBERO-Spatial, Object, Goal, and Long (LIBERO-10). Each suite has 10 tasks and tests understanding of spatial relations, identification of the target object, understanding of the goal, and long-horizon manipulation with multiple steps, respectively.

\textbf{Data and input/output.} Each task has 50 successful demonstrations collected by human teleoperation. The 50 demonstrations of each task are split by episode into 45 for training and 5 for validation. The observation consists of two $224\times224$ RGB images from the external and wrist views, and a 10-dimensional end-effector state $s_t$ made of the 3D end-effector position, the 6D rotation representation, and the 1D gripper finger opening. The action is 7-dimensional: 6D delta translation and rotation of the end-effector and a 1D gripper command. The future end-effector trajectory has $K=16$ points for each view, and the action chunk length is $H=16$ steps. As data augmentation, random translation of up to $\pm8$ pixels and color jitter (brightness, contrast, and saturation) are applied. When the image is translated, the image coordinates of the trajectory labels are shifted by the same amount to keep the image and the labels geometrically consistent.

\textbf{Training.} We use the Adam optimizer, a batch size of 256, a learning rate of $4\times10^{-4}$ ($2\times10^{-5}$ for the pretrained vision backbone), gradient clipping with a maximum norm of 1.0, and bfloat16 mixed precision. The learning rate is constant. An exponential moving average (EMA, decay 0.995) of the parameters is updated every step, and the EMA parameters are used for evaluation. The loss weights in Eq.~\eqref{eq:loss} are $\lambda_{\mathrm{KP}}=20$ and $\lambda_{\mathrm{ORI}}=10$. The number of training epochs is 1,000 for Spatial, Object, and Goal, and 500 for Long. All models are trained on a single NVIDIA RTX PRO 6000 Blackwell GPU, and training takes 5 to 11 hours per model.

\textbf{Evaluation.} For each task, we use the 50 official initial states provided by LIBERO, giving 500 trials per suite and 2,000 trials over the four suites. After setting the initial state, 20 no-op steps are inserted to let the objects settle, and then the policy starts. The maximum number of steps is 330 for Spatial and Object, 350 for Goal, and 570 for Long. Success is judged by the LIBERO success conditions.

\textbf{LIBERO-Plus.} LIBERO-Plus adds seven types of perturbations to each standard LIBERO task: camera viewpoint, robot initial state, language instruction, lighting, background texture, sensor noise, and object layout (added distractor objects and displaced target objects). We use the policy trained on standard LIBERO as is, without extra training on perturbed data. Following the official protocol, each perturbed task is run once from one official initial state. The evaluation covers 10,030 tasks: 2,402 for Spatial, 2,518 for Object, 2,591 for Goal, and 2,519 for Long. The number of trials per perturbation is 1,564 for Cam., 1,515 for Robot, 1,537 for Lang., 1,142 for Light, 1,202 for Bk., 1,566 for Noise, and 1,504 for Layout. Paraphrased language instructions are encoded at run time by the same frozen SigLIP text encoder as in training.

\textbf{Ablation conditions.}
We separate the contributions of the two components of FRAM: (i) the reference positions given by the future trajectory, and (ii) the local visual features read from those positions.
All conditions use the same data, training settings, and checkpoint selection rule.
\minusfeat{} keeps the predicted trajectory but removes the local visual features $f_{\mathrm{local}}^v$ read from the trajectory positions.
In this condition, the policy knows where to look from the future trajectory, but cannot use the local image information that represents the pose and state of the object at that position. Comparison with \minusfeat{} therefore isolates the contribution of reading local visual features from the trajectory positions, apart from the effect of the predicted trajectory itself.
\minustraj{} removes the trajectory predictor, and with it the local visual features and the trajectory tokens.
In this condition, no explicit information about where to look in the image is given, and the position and state of the target object are represented only implicitly in the global visual features. Comparison with \minustraj{} therefore shows the contribution of using the future trajectory as spatial reference positions.
In particular, we evaluate how much explicit reference positions contribute to performance and to robustness against position changes when the object layout, camera viewpoint, and robot initial pose change.

\begin{table}[t]
\caption{Success rates (\%) on standard LIBERO}
\label{tab:standard}
\centering
\begingroup
\fontsize{8}{10}\selectfont
\setlength{\tabcolsep}{2pt}
\renewcommand{\arraystretch}{1.12}

\begin{tabular*}{\columnwidth}{@{\extracolsep{\fill}}lrrrrrr@{}}
\toprule
Model & Size & Spatial & Object & Goal & Long & Ave. \\
\midrule
OpenVLA-OFT~\cite{oft} & 7.71B & 97.6 & 98.4 & 97.9 & 94.5 & 97.1 \\
FLOWER~\cite{flower} & 950M & 97.5 & 99.1 & 96.1 & 94.9 & 96.9 \\
$\pi_0$~\cite{pi0} & 3.3B & 96.8 & 98.8 & 95.8 & 85.2 & 94.2 \\
\textbf{FRAM (Our)} & \textbf{138M} & \textbf{92.2} & \textbf{96.2} & \textbf{93.8} & \textbf{86.4} & \textbf{92.2} \\
SmolVLA~\cite{smolvla} & 450M & 90.0 & 96.0 & 92.0 & 71.0 & 87.3 \\
$\pi_0$-FAST~\cite{fast} & 3B & 96.4 & 96.8 & 88.6 & 60.2 & 85.5 \\
SmolVLA~\cite{smolvla} & 240M & 87.0 & 93.0 & 88.0 & 63.0 & 82.8 \\
OpenVLA~\cite{openvla} & 7.54B & 84.7 & 88.4 & 79.2 & 53.7 & 76.5 \\
Octo~\cite{octo} & 204M & 78.9 & 85.7 & 84.6 & 51.1 & 75.1 \\
\cmidrule{1-7}
FRAM -Feat. & 137M & 88.0 & 96.8 & 92.0 & 76.6 & 88.4 \\
FRAM -Traj. & 129M & 83.6 & 94.8 & 93.6 & 78.8 & 87.7 \\
\bottomrule
\end{tabular*}
\endgroup
\end{table}

\subsection{Success Rates on Standard Tasks}
\label{ssec:sim_results}

Table~\ref{tab:standard} shows the task success rates on the four standard LIBERO suites. FRAM reaches 92.2\%, 96.2\%, 93.8\%, and 86.4\% on Spatial, Object, Goal, and Long, respectively, with an average of 92.2\% over the four suites. While the success rate is above 92\% on Spatial, Object, and Goal, it drops to 86.4\% on Long, which involves multiple steps, leaving room for improvement in long-horizon manipulation.

We compare with small models and large model separately.
Among the small models, SmolVLA (450M, 240M)~\cite{smolvla} and Octo~\cite{octo} reach average success rates of 87.3\%, 82.8\%, and 75.1\%, respectively, while FRAM reaches 92.2\%, which is 4.9--17.1 points higher with a smaller model. This shows that FRAM keeps high performance with a small model not by simply shrinking the model, but by structuring visual information with the future trajectory.

Among the large models, OpenVLA-OFT~\cite{oft} and FLOWER~\cite{flower} outperform FRAM with 97.1\% and 96.9\%, but the gap is only about 5 points. FRAM is on par with $\pi_0$~\cite{pi0} at 94.2\%, and outperforms $\pi_0$-FAST~\cite{fast} at 85.5\% and OpenVLA ~\cite{openvla} at 76.5\%. Therefore, although FRAM is the smallest model in the table, it clearly outperforms the small models and matches large models that are several to tens of times larger.

\begin{table}[t]
\caption{Success rates (\%) per perturbation on LIBERO-Plus}
\label{tab:plus}
\begingroup
\fontsize{8}{10}\selectfont
\setlength{\tabcolsep}{1pt}
\renewcommand{\arraystretch}{1.12}

\begin{tabular*}{86mm}{@{\extracolsep{\fill}}lrrrrrrrr@{}}
\toprule
Model & Cam. & Robot & Lang. & Light & Bk. & Noise & Layout & Ave. \\
\midrule
OpenVLA-OFT~\cite{oft}
& 56.4 & 31.9 & 79.5 & 88.7 & 93.3 & 75.8 & 74.2 & 69.6 \\
\textbf{FRAM (Ours)}
& \textbf{68.6} & \textbf{40.2} & \textbf{70.7} & \textbf{87.1}
& \textbf{68.5} & \textbf{72.4} & \textbf{68.3} & \textbf{67.3} \\
$\pi_0$-FAST~\cite{fast}
& 65.1 & 21.6 & 61.0 & 73.2 & 73.2 & 74.4 & 68.8 & 61.6 \\
$\pi_0$~\cite{pi0}
& 13.8 & 6.0 & 58.8 & 85.0 & 81.4 & 79.0 & 68.9 & 53.6 \\
OpenVLA~\cite{openvla}
& 0.8 & 3.5 & 23.0 & 8.1 & 34.8 & 15.2 & 28.5 & 15.6 \\ 
\cmidrule{1-9}
FRAM -Feat.
& 61.6 & 33.1 & 55.0 & 81.4
& 67.9 & 57.1 & 56.4 & 57.8 \\
FRAM -Traj.
& 52.7 & 28.5 & 47.8 & 80.6
& 56.2 & 46.9 & 51.7 & 50.8 \\
\bottomrule
\end{tabular*}

\endgroup
\end{table}

\begin{figure*}[t]
\centering
\includegraphics[width=0.98\textwidth]{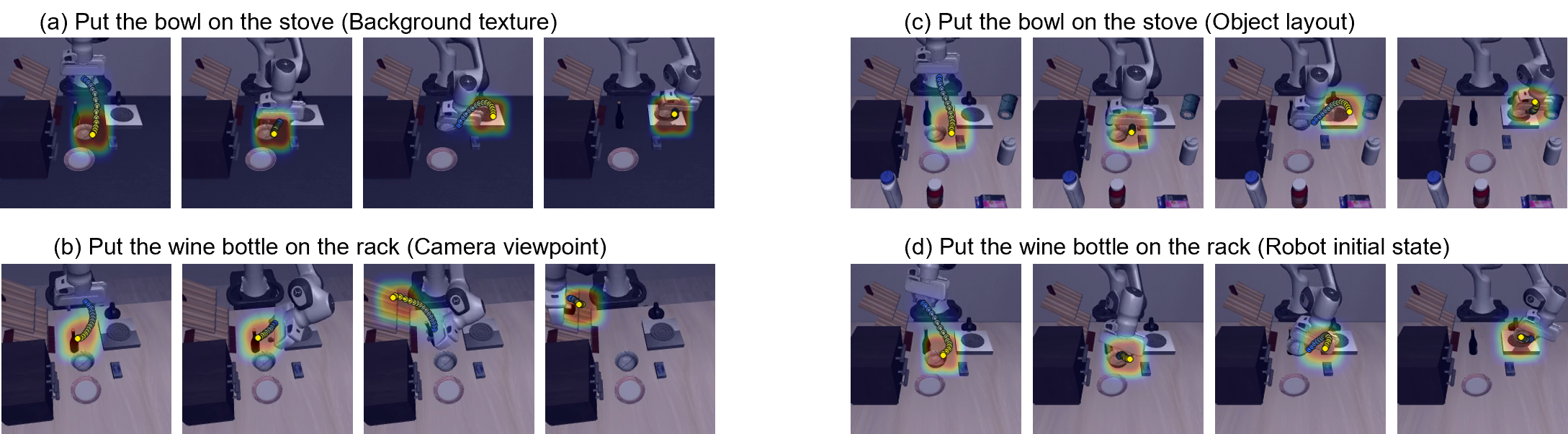}
\caption{Examples of FRAM under LIBERO-Plus perturbations. (a) Background texture, (b) camera viewpoint, (c) object layout, and (d) robot initial pose are changed. The dots show the predicted trajectory, the end dot shows the target location, and the heatmap shows the local features.}
\label{fig:plus_results}
\end{figure*}

\subsection{Robustness to Environment Changes}
\label{ssec:robustness}

Table~\ref{tab:plus} shows the results on LIBERO-Plus.
FRAM reaches an average success rate of 67.3\% over all perturbations.
This is close to the 69.6\% of OpenVLA-OFT,
and above the 61.6\% of $\pi_0$-FAST, 53.6\% of $\pi_0$, and 15.6\% of OpenVLA.
For camera viewpoint and robot initial state in particular,
FRAM reaches 68.6\% and 40.2\%,
which is 12.2 and 8.3 points above OpenVLA-OFT.
On the other hand, on background it reaches 68.5\%,
far below the 93.3\% of OpenVLA-OFT.
In summary, FRAM is especially robust to changes in viewpoint and robot initial pose,
while there is room for improvement on background changes.

Fig.~\ref{fig:plus_results} shows examples under typical perturbations.
Even when the background texture or object layout changes,
the predicted trajectory and the heatmap follow the target object and the target location.
When the camera viewpoint or the robot initial state changes,
a future trajectory toward the target is still generated step by step from the current observation.
These examples qualitatively show the robustness of FRAM to environment changes seen in Table~\ref{tab:plus}.

\begin{figure*}[t]
\centering
\includegraphics[width=0.98\textwidth]{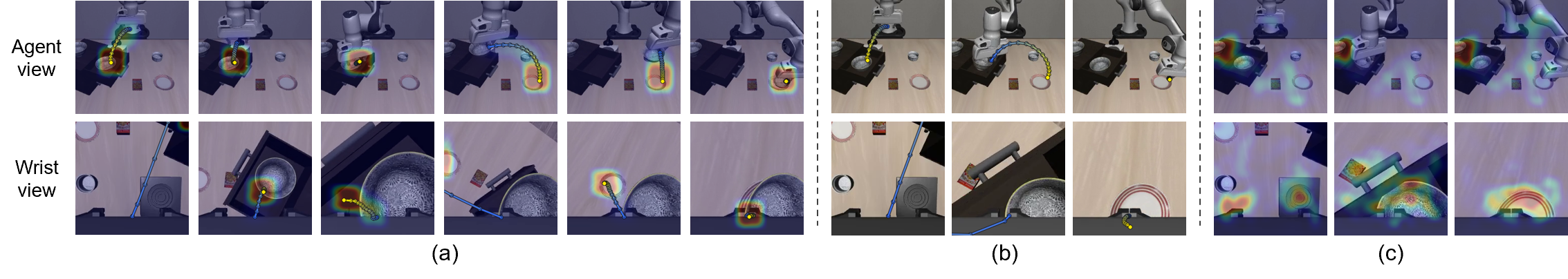}
\caption{Examples from the same initial state. (a) FRAM, (b) \minusfeat{}, and (c) \minustraj{}. The top row is the external view and the bottom row is the wrist view. The dots show the predicted keypoint trajectory and the heatmap shows the visual features used for action generation. (b) predicts the trajectory but uses no image features, so no heatmap is shown. (c) uses the global ResNet feature map, shown as a heatmap over the whole image, but predicts no trajectory.}
\label{fig:ablation}
\end{figure*}

\subsection{Ablation}
\label{ssec:ablation}

To verify the effectiveness of the proposed future trajectory and local visual features,
we compare with \minusfeat{} and \minustraj{}, each of which removes one component.
As shown in Table~\ref{tab:standard},
the average success rate on standard LIBERO drops from 92.2\% for FRAM
to 88.4\% for \minusfeat{} and 87.7\% for \minustraj{}.
In particular, it drops from 92.2\% to 88.0\% and 83.6\% on Spatial,
and from 86.4\% to 76.6\% and 78.8\% on Long.
These results confirm that the reference positions given by the future trajectory,
and the local visual features read from those positions,
both contribute to high performance.

On LIBERO-Plus, which includes environment changes, the gap is even larger.
As shown in Table~\ref{tab:plus},
the average success rate drops from 67.3\% for FRAM
to 57.8\% for \minusfeat{} and 50.8\% for \minustraj{}.
Both \minusfeat{} and \minustraj{} are below FRAM on all seven perturbations.
Therefore, the combination of the proposed future trajectory and local features
is effective not only for performance in the standard environment
but also for robustness to environment changes.

Fig.~\ref{fig:ablation} shows examples from the same initial state.
In FRAM, the predicted trajectory is updated from the object to grasp to the placement location as the task progresses,
local features are read from around those positions, and the task succeeds.
In contrast, \minusfeat{} and \minustraj{} fail the task
because part of the proposed representation is removed.
These examples also show the effectiveness of combining spatial reference by the future trajectory
with local feature extraction from those positions.

\section{Real-Robot Experiments}
\subsection{Setup}
\label{ssec:real_setup}

\textbf{Robot and task.}
As an example of application to a real environment, we built a cup stacking task with a dual-arm UR5e.
Each arm has 6 DoF and a parallel gripper.
Four cups of different sizes and colors are placed on a table.
The largest, indigo cup is the base, and
the orange, blue, and yellow cups are stacked on it in this order.
Only two wrist cameras mounted on the left and right wrists
(GoPro HERO13 Black, wide-angle mode) are used as observations, without any external camera.
The central square region of each image is cropped and resized to $224\times224$ pixels.

\textbf{Data collection.}
We collected 62 episodes of demonstrations
(about 44 minutes, about 30~Hz, 79,669 frames in total)
by leader-follower teleoperation.
In each episode, the three cups are grasped in order and stacked on the base cup.
The operator chose which arm to use according to the cup layout:
44 episodes use both arms, 8 use only the left arm, and 10 use only the right arm.
The policy therefore has to decide from the observation not only the next target object,
but also which arm to use and when to move it.

\begin{figure*}[t]
\centering
\includegraphics[width=0.98\textwidth]{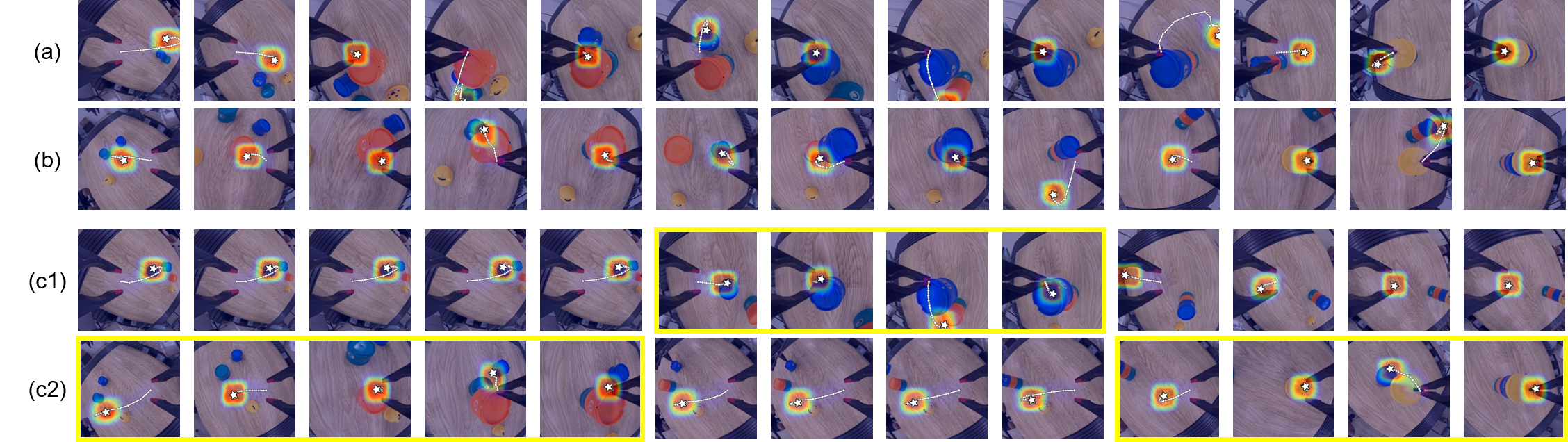}
\caption{Cup stacking with a dual-arm UR5e. (a) Left arm only, (b) right arm only, and (c) both arms (c1: left wrist, c2: right wrist). The white line is the predicted trajectory, the star is its end point, and the yellow frame marks the arm in motion.}
\label{fig:ur_stacking}
\end{figure*}

\subsection{Results}
\label{ssec:real_results}

Fig.~\ref{fig:ur_stacking} shows autonomous cup stacking by the trained policy.
In (a), the three target cups are placed mainly on the left arm side,
and the policy performed all operations with the left arm.
The predicted trajectory on the left wrist image first heads to the orange cup,
and after grasping, its end point moves to the indigo base cup.
After placing, the reference moves to the next blue cup and then to the yellow cup,
stacking them in the order orange, blue, and yellow.
In (b), the three cups are handled only by the right arm.
On the right wrist image as well, the predicted trajectory and its end point
switch between the next object to grasp and the placement location as the task progresses,
and the same trajectory prediction and local feature reading are observed even when a different arm is used.

In (c), the cups are spread over both sides,
and the policy executes the task while switching between the left and right arms.
The yellow frame marks the wrist image of the arm that is actually moving at each time.
Early on, the right arm stacks the orange cup on the base,
then the left arm stacks the blue cup, and finally the right arm stacks the yellow cup.
In this way, a single policy generates the sequence right arm, left arm, right arm
according to the cup layout and the task progress.

Of particular interest is the predicted trajectory of the waiting arm, which has no yellow frame.
Early in (c1), while the right arm handles the orange cup,
the predicted trajectory and reading positions of the left arm already appear near the blue cup that the left arm will handle next.
However, the left arm does not move at this point.
It moves on to grasping and stacking the blue cup
only after the right arm has finished placing the orange cup.
After that, when no target remains for the left arm,
the predicted trajectory stays near the end-effector and does not extend toward any cup.
Similarly, in (c2), while the left arm handles the blue cup,
the predicted trajectory of the right arm appears near the yellow cup that comes next,
and the right arm moves on to grasping the yellow cup after the left arm finishes.

These results confirm that the predicted trajectory and local feature reading positions of FRAM
change not only with the current object to grasp and the placement location, but also with the subsequent target.
In addition, even when the predicted trajectory of the waiting arm points to a future target,
that arm does not move right away; it starts after the other arm completes its preceding operation.
In summary, FRAM generates a sequence of cup stacking operations in real dual-arm manipulation,
including choosing, waiting, and switching between the left and right arms,
while updating the future trajectory step by step according to the target object and the placement location.

\section{Conclusion}
In this work, we proposed FRAM, a small robot policy that explicitly links future motion to the visual information related to the action. FRAM predicts the future end-effector trajectory and uses its image coordinates as spatial pointers to read local visual features from the current image, structuring the information needed for manipulation as the reference position (Where), the current visual state (What), and the future motion (Future). Trajectory labels are generated automatically from demonstrations and camera geometry, so no manual annotation is needed.

On LIBERO, with 138.7M parameters including a frozen language encoder, FRAM reached an average success rate of 92.2\% over the four standard suites, outperforming the small models we compared with and matching models several to tens of times larger. Without extra training on perturbed environments, it reached 67.3\% on LIBERO-Plus and was especially robust to changes in camera viewpoint and robot initial pose. Ablations confirmed that both the reference positions given by the future trajectory and the local visual features read from those positions contribute to performance and robustness. In real cup stacking with a dual-arm UR5e, using only wrist cameras, the predicted trajectory was updated step by step according to the target object, the placement location, and the subsequent target, and a single policy generated a sequence of operations including choosing, waiting, and switching between the left and right arms.

As future work, the success rate on LIBERO-Long, which involves multiple steps, is lower than on the other suites, and performance in long-horizon manipulation needs to improve.
Also, since the real-robot evaluation in this work is limited to cup stacking, we will apply FRAM to a wider range of real tasks with different objects, task contents, environment conditions, and robot configurations to further verify its generality and effectiveness in real environments.

\bibliographystyle{IEEEtran}
\bibliography{ref}

@inproceedings{spatialae,
 author={Finn, Chelsea and Tan, Xin Yu and Duan, Yan and Darrell, Trevor and Levine, Sergey and Abbeel, Pieter},
 title={Deep Spatial Autoencoders for Visuomotor Learning},
 booktitle={Proc. IEEE International Conference on Robotics and Automation (ICRA)},
 year={2016}
}

@inproceedings{atm,
 author={Wen, Chuan and Lin, Xingyu and So, John and Chen, Kai and Dou, Qi and Gao, Yang and Abbeel, Pieter},
 title={Any-point Trajectory Modeling for Policy Learning},
 booktitle={Proc. Robotics: Science and Systems (RSS)}, year={2024},
 doi={10.15607/RSS.2024.XX.092}
}

@inproceedings{mtpi,
 author={Ren, Juntao and Sundaresan, Priya and Sadigh, Dorsa and Choudhury, Sanjiban and Bohg, Jeannette},
 title={Motion Tracks: A Unified Representation for Human-Robot Transfer in Few-Shot Imitation Learning},
 booktitle={Proc. IEEE International Conference on Robotics and Automation (ICRA)},
 year={2025}
}

@misc{rttrajectory,
 author={Gu, Jiayuan and Kirmani, Sean and Wohlhart, Paul and Lu, Yao and Gonzalez Arenas, Montserrat and Rao, Kanishka and others},
 title={{RT-Trajectory}: Robotic Task Generalization via Hindsight Trajectory Sketches},
 howpublished={arXiv:2311.01977}, year={2023}
}

@inproceedings{flowmatching,
 author={Lipman, Yaron and Chen, Ricky T. Q. and Ben-Hamu, Heli and Nickel, Maximilian and Le, Matt},
 title={Flow Matching for Generative Modeling},
 booktitle={Proc. International Conference on Learning Representations (ICLR)},
 year={2023}
}

@inproceedings{rot6d,
 author={Zhou, Yi and Barnes, Connelly and Lu, Jingwan and Yang, Jimei and Li, Hao},
 title={On the Continuity of Rotation Representations in Neural Networks},
 booktitle={Proc. IEEE/CVF Conference on Computer Vision and Pattern Recognition (CVPR)},
 pages={5745--5753}, year={2019}
}

@inproceedings{siglip,
 author={Zhai, Xiaohua and Mustafa, Basil and Kolesnikov, Alexander and Beyer, Lucas},
 title={Sigmoid Loss for Language Image Pre-Training},
 booktitle={Proc. IEEE/CVF International Conference on Computer Vision (ICCV)},
 pages={11975--11986}, year={2023}
}

@inproceedings{film,
 author={Perez, Ethan and Strub, Florian and de Vries, Harm and Dumoulin, Vincent and Courville, Aaron},
 title={{FiLM}: Visual Reasoning with a General Conditioning Layer},
 booktitle={Proc. AAAI Conference on Artificial Intelligence},
 volume={32}, number={1}, year={2018},
 doi={10.1609/aaai.v32i1.11671}
}

@inproceedings{libero,
 author={Liu, Bo and Zhu, Yifeng and Gao, Chongkai and Feng, Yihao and Liu, Qiang and Zhu, Yuke and Stone, Peter},
 title={{LIBERO}: Benchmarking Knowledge Transfer for Lifelong Robot Learning},
 booktitle={Advances in Neural Information Processing Systems},
 volume={36}, year={2023}
}

@inproceedings{liberoplus,
 author={Fei, Senyu and Wang, Siyin and Shi, Junhao and Dai, Zihao and Cai, Jikun and Qian, Pengfang and Ji, Li and He, Xinzhe and Zhang, Shiduo and Fei, Zhaoye and Fu, Jinlan and Gong, Jingjing and Qiu, Xipeng},
 title={{LIBERO-Plus}: A Progressive Robustness Benchmark for Visual-Language-Action Models},
 booktitle={Proc. IEEE/CVF Conference on Computer Vision and Pattern Recognition (CVPR)},
 year={2026}
}

@misc{openvla,
 author={Kim, Moo Jin and Pertsch, Karl and Karamcheti, Siddharth and Xiao, Ted and Balakrishna, Ashwin and Nair, Suraj and others},
 title={{OpenVLA}: An Open-Source Vision-Language-Action Model},
 howpublished={arXiv:2406.09246}, year={2024}
}

@misc{pi0,
 author={Black, Kevin and Brown, Noah and Driess, Danny and Esmail, Adnan and Equi, Michael and Finn, Chelsea and others},
 title={{$\pi_0$}: A Vision-Language-Action Flow Model for General Robot Control},
 howpublished={arXiv:2410.24164}, year={2024}
}

@misc{oft,
 author={Kim, Moo Jin and Finn, Chelsea and Liang, Percy},
 title={Fine-Tuning Vision-Language-Action Models: Optimizing Speed and Success},
 howpublished={arXiv:2502.19645}, year={2025}
}

@misc{smolvla,
 author={Shukor, Mustafa and Aubakirova, Dana and Capuano, Francesco and Kooijmans, Pepijn and Palma, Steven and Zouitine, Adil and others},
 title={{SmolVLA}: A Vision-Language-Action Model for Affordable and Efficient Robotics},
 howpublished={arXiv:2506.01844}, year={2025}
}

@misc{fast,
 author={Pertsch, Karl and Stachowicz, Kyle and Ichter, Brian and Driess, Danny and Nair, Suraj and Vuong, Quan and Mees, Oier and Finn, Chelsea and Levine, Sergey},
 title={{FAST}: Efficient Action Tokenization for Vision-Language-Action Models},
 howpublished={arXiv:2501.09747}, year={2025}
}

@inproceedings{octo,
 author={{Octo Model Team} and Ghosh, Dibya and Walke, Homer and Pertsch, Karl and Black, Kevin and Mees, Oier and others},
 title={Octo: An Open-Source Generalist Robot Policy},
 booktitle={Proc. Robotics: Science and Systems (RSS)}, year={2024}
}

@misc{ript,
 author={Tan, Shuhan and Dou, Kairan and Zhao, Yue and Kr{\"a}henb{\"u}hl, Philipp},
 title={{Interactive Post-Training for Vision-Language-Action Models}},
 howpublished={arXiv:2505.17016}, year={2025}
}

@misc{thinkact,
 author={Huang, Chi-Pin and Wu, Yueh-Hua and Chen, Min-Hung and Wang, Yu-Chiang Frank and Yang, Fu-En},
 title={{ThinkAct: Vision-Language-Action Reasoning via Reinforced Visual Latent Planning}},
 howpublished={arXiv:2507.16815}, year={2025}
}

@misc{cotvla,
 author={Zhao, Qingqing and Lu, Yao and Kim, Moo Jin and Fu, Zipeng and Zhang, Zhuoyang and Wu, Yecheng and Li, Zhaoshuo and Ma, Qianli and Han, Song and Finn, Chelsea and Handa, Ankur and Liu, Ming-Yu and Xiang, Donglai and Wetzstein, Gordon and Lin, Tsung-Yi},
 title={{CoT-VLA: Visual Chain-of-Thought Reasoning for Vision-Language-Action Models}},
 howpublished={arXiv:2503.22020}, year={2025}
}

@inproceedings{flower,
 author={Reuss, Moritz and Zhou, Hongyi and R{\"u}hle, Marcel and Ya{\u g}murlu, {\"O}mer Erdin{\c c} and Otto, Fabian and Lioutikov, Rudolf},
 title={{FLOWER}: Democratizing Generalist Robot Policies with Efficient Vision-Language-Flow Models},
 booktitle={Proceedings of the 9th Conference on Robot Learning},
 series={Proceedings of Machine Learning Research}, volume={305}, pages={3736--3761}, year={2025}
}

@article{land1999,
 author={Land, Michael F. and Mennie, Neil and Rusted, Jennifer},
 title={The Roles of Vision and Eye Movements in the Control of Activities of Daily Living},
 journal={Perception}, volume={28}, number={11}, pages={1311--1328}, year={1999},
 doi={10.1068/p2935}
}

@article{johansson2001,
 author={Johansson, Roland S. and Westling, G{\"o}ran and B{\"a}ckstr{\"o}m, Anders and Flanagan, J. Randall},
 title={Eye--Hand Coordination in Object Manipulation},
 journal={The Journal of Neuroscience}, volume={21}, number={17}, pages={6917--6932}, year={2001},
 doi={10.1523/JNEUROSCI.21-17-06917.2001}
}

\end{document}